\documentclass[conference]{IEEEtran}

\ifCLASSINFOpdf

\else
 
\fi

\usepackage{multirow}
\usepackage{array}
\usepackage{graphicx}
\usepackage{algorithm}
\usepackage{color}
\usepackage[dvipsnames]{xcolor}
\usepackage[noend]{algpseudocode}
\usepackage{url}
\usepackage{amsmath, amssymb}
\usepackage{bbm}
\usepackage{booktabs}
\begin{document}

\title{Spatially-Adaptive Conformal Graph Transformer for Indoor Localization in Wi-Fi Driven Networks}


\author{Ayesh~Abu~Lehyeh\IEEEauthorrefmark{1}, Anastassia~Gharib\IEEEauthorrefmark{4}, Safwan~Wshah\IEEEauthorrefmark{1}\\\IEEEauthorrefmark{1}Department of Computer Science, The University of Vermont, Burlington, VT 05405, USA\\\IEEEauthorrefmark{4}Department of Computer Science \& Engineering, American University of Sharjah, Sharjah 26666, UAE\\Emails: Ayesh.Abulehyeh@uvm.edu; AGharib@aus.edu; Safwan.Wshah@uvm.edu}


\maketitle

\begin{abstract}
Indoor localization is a critical enabler for a wide range of location-based services in smart environments, including navigation, asset tracking, and safety-critical applications. 
Recent graph-based models leverage spatial relationships between Wireless Fidelity (Wi-Fi) Access Points (APs) and devices, offering finer localization granularity, 
but 
fall short in quantifying prediction uncertainty, a key requirement for real-world deployment. In this paper, we propose Spatially-Adaptive Conformal Graph Transformer (SAC-GT), a 
framework for accurate and reliable indoor localization. SAC-GT integrates a Graph Transformer (GT) model that captures  network's spatial topology and  signal strength 
dynamics, with a novel Spatially-Adaptive Conformal Prediction (SACP) method that provides region-specific uncertainty estimates. This allows SAC-GT to produce not only precise two-dimensional (2D) location predictions but also statistically valid confidence regions tailored to varying environmental conditions. Extensive evaluations on a large-scale real-world dataset demonstrate that the proposed SAC-GT solution achieves state-of-the-art localization accuracy while delivering robust and spatially adaptive reliability guarantees. 

\textit{Index Terms}---Conformal Prediction (CP), Graph Transformer (GT), indoor localization, Received Signal Strength Indicator (RSSI), Wireless Fidelity (Wi-Fi), wireless networks.
\end{abstract}

\IEEEpeerreviewmaketitle

\section{Introduction}
Accurate and reliable indoor localization is a cornerstone of modern smart environments~\cite{CHO2025100164}. They enable a wide range of location-based services (LBS) in complex indoor venues (such as airports, hospitals, and shopping malls) with applications ranging from indoor navigation and asset tracking to safety-critical functions~\cite{8409950}. 
These critical applications demand solutions that are not only precise but also highly reliable. However, achieving this remains a significant challenge. This is because, unlike open outdoor environments, where the Global Positioning System (GPS) performs effectively, indoor environments typically obstruct the line-of-sight (LoS) required for satellite-based positioning~\cite{8409950}. As a result, GPS and other satellite navigation systems become either unreliable or entirely unavailable~indoors. 

To overcome these limitations, many systems leverage existing wireless infrastructure, particularly Wi-Fi Access Points (APs)~\cite{cnnloc,9647082}.
These APs act as signal transmitters. By analyzing the characteristics of the signals received from them (such as signal strength or phase), it is possible to estimate a device's location indoors.
While some methods utilize Channel State Information (CSI), which provides fine-grained amplitude and phase data for high accuracy, Wi-Fi Received Signal Strength Indicator (RSSI) remains a more practical, cost-effective, and universally available solution~\cite{cnnloc}.
Historically, RSSI-based localization has been approached using techniques like multilateration and fingerprinting. Multilateration, attempts to calculate a position by converting signal strength into distances to multiple known APs~\cite{Ubigtloc}. However, this method struggles in complex indoor environments where signal attenuation and multipath effects make the relationship between RSSI and distance highly unreliable. In contrast, fingerprinting is a more common approach in indoor environments. This involves creating a detailed radio map of RSSI values at known locations (reference points) in an offline phase and then matching a user's real-time RSSI scan to this map~\cite{Ubigtloc}. While conceptually simple, traditional fingerprint-matching algorithms exhibit significant limitations in high-dimensional pattern analysis and are not adaptive to the dynamic signal fluctuations common in real-world environments, which hinders their performance.

To address these limitations, deep learning models emerged as a promising alternative~\cite{cnnloc,ahmed_unified_2024}.
They 
automatically extract salient representations from high-dimensional and noisy signal data, effectively modeling the complex non-linear relationships that traditional methods struggle to capture. For instance, the work in~\cite{cnnloc} employs RSSI and CSI features with Multi-Layer Perception (MLP) and one-dimensional Convolutional Neural Network (CNN) architectures, showing 
improved accuracy with manageable complexity. Another work in~\cite{ahmed_unified_2024} proposes a Unified deep transfer learning model based on MLP (U-MLP) to handle localization in both indoor and outdoor settings. U-MLP 
uses encoder-based transfer learning 
for training purposes. However, these conventional deep learning models treat the signals as a flat vector, failing to account for the underlying graph structure of the wireless network.

In contrast to conventional deep learning models, graph-based deep learning models 
can capture the structural relationships between APs and mobile users. Some works in this area formulated the problem as a classification task. For instance, the works in \cite{9647082} and \cite{10.1007/978-3-031-49601-1_11} utilized Graph Convolutional Networks (GCNs) to classify the user's location to regions, achieving high region-level accuracy. However, such classification-based approaches are inherently limited in precision, as their accuracy is constrained by the granularity of the predefined location labels.
To achieve fine-grained localization accuracy, subsequent research has focused on graph-based regression. The work in \cite{9964074} used GCNs with CSI measurements to achieve a high localization accuracy of $1.19$~m. Following, GC-Loc  was proposed, utilizing a Graph Attention Network (GAT) with residual connections and ensemble learning for collaborative indoor localization in~\cite{10.1145/3569495}. In addition, GLoc was proposed in~\cite{10843349}. GLoc presented a GAT-based model that uses CSI features extracted via a convolutional network to learn complex signal relationships between AP and user nodes. While these regression-based graph models provide accurate location estimates, they lack a rigorous framework for quantifying the reliability and uncertainty of their predictions, which is essential for many real-world deployments.
Previously, we proposed a unified Bidirectional Long Short-Term Memory (BiLSTM)-Graph Transformer (GT) based localization   framework (UBiGTLoc)~\cite{Ubigtloc}. Designed for outdoor environments,  UBiGTLoc considered both temporal and spatial features in anchor-based and anchor-free scenarios. Still, UBiGTLoc  lacks a formal method for quantifying the uncertainty of its predictions,   critical  for indoor deployments.

Recent efforts to address uncertainty in wireless localization have explored various techniques. 
Monte Carlo (MC) dropout, for instance, has been employed as an efficient Bayesian approximation technique in CNN-based models in~\cite{mcdropout}. By running multiple stochastic forward passes, it enables uncertainty estimation 
when applied to mmWave MIMO-based localization.
While useful, MC dropout lacks the formal statistical guarantees required for safety-critical applications.
Conformal Prediction (CP) has emerged as a powerful, model-agnostic framework for providing such guarantees. It allows for converting point predictions into prediction sets with a statistically valid coverage guarantee~\cite{zhou2025conformalprediction}. The framework in~\cite{naviguncertainty} uses CP to quantify ambiguity in fingerprinting indoor localization using both CSI and Bluetooth signals. More recently, The work in~\cite{zhou2025conformalprediction} applied CP to a CNN-based classifier to provide reliable uncertainty estimates. While these methods provide a significant step towards reliability, their guarantees are typically global, applying a single uncertainty margin across the entire indoor environment. This overlooks the intuitive reality of indoor spaces where different regions (e.g., open areas vs. complex hallways) exhibit 
different signal behaviors and thus,  require a unique, spatially-adaptive uncertainty estimate.

To address these gaps, we propose Spatially-Adaptive Conformal Graph Transformer (SAC-GT), a unified framework that provides both high point localization accuracy and a novel, region-aware reliability guarantee. The contributions of this paper are as follows.
\begin{itemize}
    \item 
    We propose a robust GT model 
    that leverages the fixed spatial locations of APs and the dynamic Wi-Fi RSSI signals to learn a direct, end-to-end mapping from a graph representation to precise 2D~coordinates.
    \item 
    We introduce a novel Spatially-Adaptive Conformal Prediction (SACP) framework, which is theoretically applicable to any deep learning–based indoor localization model. The method partitions the environment into distinct spatial zones and calibrates a unique, statistically valid confidence radius for each, yielding a more granular and practically meaningful uncertainty estimation. 
    \item 
    We demonstrate the effectiveness of our SAC-GT framework through extensive experiments on the well-known SODIndoorLoc dataset, a large-scale, real-world benchmark in the field. Our approach achieves state-of-the-art performance, delivering both high accuracy and precision in point predictions while providing reliable and spatially adaptive confidence regions.
\end{itemize}

The rest of the paper is structured as follows. Sections~II~and~III introduce the system model and the problem formulation, respectively. In Section IV, the proposed SAC-GT framework is presented. Simulation results are discussed in Section~V. Finally, Section VI concludes the paper.

\section{System Model} 
\begin{figure}[t!]
\centerline{
      \includegraphics[width=0.3\textwidth]{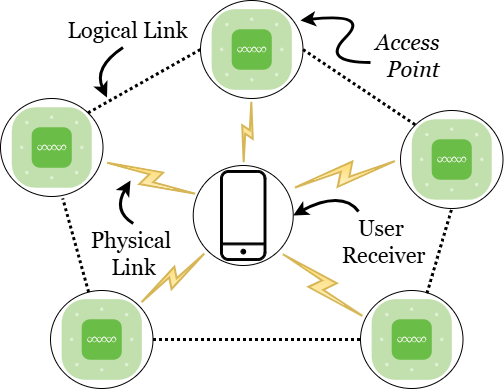} 
			}
      \caption{System model for indoor localization, showing the graph structure with a mobile user (a user receiver), APs, physical links, and logical links.}
      \label{SystemModel}
   \end{figure}

Fig.~\ref{SystemModel} shows the proposed system model for indoor localization of mobile users  based on the RSSI from  Wi-Fi APs. The environment consists of a mobile user (i.e., a user receiver) and a set of $m$ pre-installed Wi-Fi APs with known 2D coordinate locations. We model this system as a graph where the nodes represent the user and  APs. The graph contains two distinct types of connections, as shown in Fig.~\ref{SystemModel}. First, physical links are formed between the user and surrounding APs (from which the user receives detectable Wi-Fi signals). These links are dynamic, representing the real-time RSSI measurements that serve as the primary sensory input for localization. Second, logical links exist between APs that are in close physical proximity to each other. These links are static and encode the underlying spatial topology of the AP infrastructure. At any given time, the user's device performs a Wi-Fi scan, resulting in an RSSI feature vector $\hat{\mathbf{F}}_u = [\text{RSSI}_1, \text{RSSI}_2, \dots, \text{RSSI}_m]$, where each element corresponds to the RSSI from one of the $m$ APs. For each AP $i$, the features are its known and fixed physical coordinates, $\hat{\mathbf{F}}_{a_i} = [x_{a_i}, y_{a_i}]$. Thus, a feature matrix $\mathbf{F}_{a} = [\hat{\textbf{F}}_{a_1}; \hat{\textbf{F}}_{a_2}; \dots \hat{\textbf{F}}_{a_m}]$ is formed, which represents the location of all $m$ APs. 
Such feature representation, combined with the graph's link structure, forms the complete input to our localization framework.

\section{Problem Formulation} 
We formulate Wi-Fi-based indoor localization as a graph-based problem. 
For each user's RSSI features (i.e., $\hat{\mathbf{F}}_u$), we construct a unique graph $G = (\mathcal{V}, \mathcal{E})$, where $\mathcal{V}$ is the set of nodes and $\mathcal{E}$ is the set of edges. The node set $\mathcal{V}$ is the union of the user node, $v_u$, and the set of $m$ AP nodes, $\mathcal{V}_{a}$. The connectivity of this graph, which captures both the static AP topology (i.e., logical links) and the dynamic user-AP connections (i.e., physical links), is encoded in an adjacency matrix $\mathbf{A}$. An entry \(A_{ij}\) of the adjacency matrix~\(\mathbf{A}\) is defined based on the node types in the following way: 
\begin{equation}
A_{ij} = 
\begin{cases}
    \mathbbm{1}(d_{ij} \le d_{p}), & \text{if } i,j \in \mathcal{V}_{a} \\
    \mathbbm{1}(\text{RSSI}_{j} \ge \tau), & \text{if } i=v_u, j \in \mathcal{V}_{a} \\
    0, & \text{otherwise}
\end{cases}
\end{equation}
\noindent where $d_{ij}$ is the euclidean distance between two APs $i$ and $j$, $d_{p}$ is a physical proximity threshold for creating logical links, and $\tau$ is the minimum signal strength threshold for creating a physical link between the user and an AP (radio range within which nodes communicate). The indicator function \(\mathbbm{1}(\cdot)\) outputs one if the condition is true (i.e., indicating a connection between the nodes), and zero otherwise (i.e., indicating no direct connectivity).

The objective of the indoor localization problem is to estimate the position of a user node (i.e., $(\hat{x}, \hat{y})$) based on the graph structure and the aforementioned features of each node in the graph (i.e., $\hat{\mathbf{F}}_u$ and $\mathbf{F}_a$). 
Thus, we 
use the Mean Absolute Error (MAE) $\mathcal{L}_{MAE}$ between the user's predicted coordinates, $(\hat{x}_u, \hat{y}_u)$, and the user's true coordinates, $(x_u, y_u)$, that minimizes the prediction's localization error 
as~follows:
\begin{equation}
\mathcal{L}_{MAE} = \mathbb{E}_{((x_u, y_u), G)} \left[ |x_u - \hat{x}_u| + |y_u - \hat{y}_u| \right],
\label{eq:objective}
\end{equation}
\noindent where the expectation $\mathbb{E}$ denotes the mean taken over all graph samples. 
The goal is to train a model $\mathcal{M}$ that produces predicted positions as close as possible to the true positions. The following section details our proposed SAC-GT framework. 

\section{A Spatially-Adaptive Conformal Graph Transformer Model for Indoor Localization}
\begin{figure}[t!]
\centerline{\includegraphics[width=0.5\textwidth]{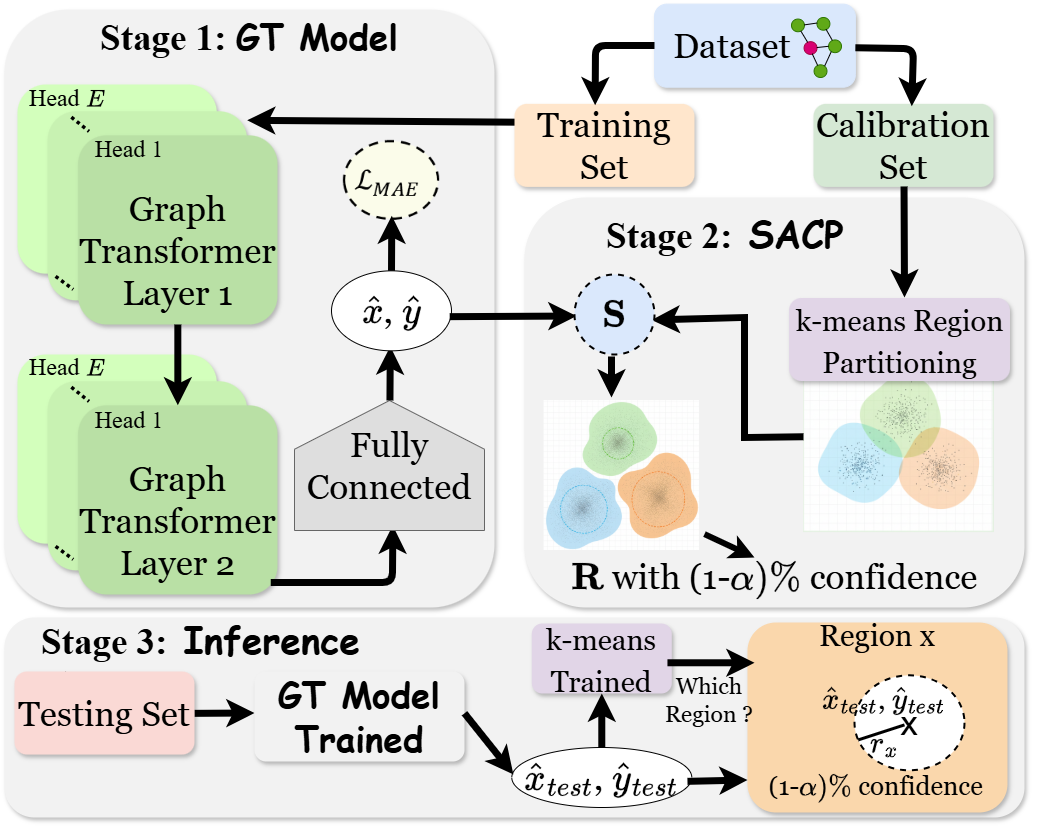}} 
      \caption{The proposed Spatially-Adaptive Conformal Graph Transformer (SAC-GT)~framework.}
      \label{model}
\end{figure}

We propose SAC-GT, a Spatially-Adaptive Conformal Graph Transformer framework for robust Wi-Fi-based indoor localization.  SAC-GT provides both accurate point predictions and reliable, adaptive confidence predictions. This adaptive approach is critical for indoor environments, where obstacles and multi-path effects cause signal behavior to vary significantly between different areas. As illustrated in Fig.~\ref{model}, the process begins by splitting the dataset into a training set and a held-out calibration set. For inference, the performance of the final, calibrated framework is then evaluated on an independent unseen testing set. Next, our framework consists of three main stages. The first stage involves training a Graph Transformer (GT) model to generate high-accuracy point predictions. In the second stage, a Spatially-Adaptive Conformal Prediction (SACP) wrapper analyzes the GT model's performance on a held-out calibration set to learn a unique, adaptive prediction radius for each geographical region. Finally, the inference stage combines the predictive power of the GT model with these learned regional radii to provide a final, statistically valid confidence for each new location estimate, ensuring the framework's output is both accurate and reliable.

\subsection{Graph Transformer (GT) Model}
The core of our SAC-GT is a GT model designed to leverage the spatial topology of the wireless indoor environment. As shown in Fig.~\ref{model}, the input to the model is a unique graph from our training set, constructed for each Wi-Fi measurement to represent each  user and their connected APs. The GT model is trained to learn a direct mapping from this graph-structured input to a final 2D coordinate prediction $(\hat{x}, \hat{y})$. Our architecture employs two stacked graph transformer layers to capture higher-order dependencies between nodes (i.e., a user and an AP). Specifically, the first layer allows the model to learn from immediate neighbors, while the second layer enables learning from the neighbors-of-neighbors, thereby expanding the model's receptive field.

For every graph transformer layer, we employ the TransformerConv operator, which combines nodes' feature aggregation with multi-head attention~\cite{shi_masked_2021}. The multi-head attention allows the model to employ~$E$ heads to capture different aspects of graph relationships, improving feature aggregation.
Thus, each layer updates each resulting node's features by aggregating information from its neighbors as~follows~\cite{shi_masked_2021}: 
\begin{equation}
\mathbf{z}_i = \mathbf{W}_1 \mathbf{\hat{F}}_i + \sum_{j \in \mathcal{N}(i)} \beta_{i,j} \mathbf{W}_2 \mathbf{\hat{F}}_j,
\end{equation}
\noindent where \( \mathbf{W}_1 \) and \( \mathbf{W}_2 \) are learnable weight matrices applied to the current feature vector (i.e., $\mathbf{\hat{F}}$) of node \( i \) and its neighbors \( j \), respectively, and \( \beta_{i,j} \) is the attention coefficient that defines the importance of node \( j \)'s information for node \( i \).
The attention coefficient \( \beta_{i,j} \) is calculated using a scaled dot-product attention mechanism 
in the following way:
\begin{equation}
\beta_{i,j} = \text{softmax} \left( \frac{\left(\mathbf{W}_3 \mathbf{\hat{F}}_i\right)^\top \left(\mathbf{W}_4 \mathbf{\hat{F}}_j\right)}{\sqrt{h}} \right),
\end{equation}
\noindent where \( \mathbf{W}_3 \) and \( \mathbf{W}_4 \) are learnable weight matrices used to project the features of nodes \( i \) and \( j \) into a shared space for comparison, and \( h \) represents the dimensionality of the feature space. The softmax function normalizes the attention scores across all neighbors of node \( i \), ensuring that the weights sum equals to one. 
This process is repeated for $E$ heads, and the results are averaged to form the final output of the layer. Consequently, the output of the GT layers is a rich, spatially-aware feature embedding, which is then passed to a fully connected layer to predict the final location ($\hat{x}, \hat{y}$) of the user node. The entire GT model is trained end-to-end to minimize MAE, i.e., $\mathcal{L}_{MAE}$ (as mentioned previously).

\subsection{Spatially-Adaptive Conformal Prediction (SACP)}

While the GT model provides accurate point predictions, it lacks a rigorous confidence measure. To address this, we build our uncertainty quantification framework using CP, a technique that converts point predictions into prediction sets with a statistically valid coverage guarantee~\cite{zhou2025conformalprediction}. Our proposed SACP extends this framework to generate reliable, adaptive confidence regions tailored to the heterogeneous nature of indoor environments.
As illustrated in Stage 2 of Fig.~\ref{model}, the SACP process operates on a held-out calibration set. First, to account for spatial variations in localization difficulty, we partition the calibration set into $k$ distinct geographical regions using K-Means clustering on the ground truth coordinates.
Next, for each calibration sample $n$, we define a nonconformity score, $s_n$, as the euclidean distance between the GT model's prediction $(\hat{x}_n, \hat{y}_n)$ and the true location $(x_n, y_n)$. This score quantifies the error for a single prediction:
\begin{equation}
    s_n = \sqrt{(\hat{x}_n - x_n)^2 + (\hat{y}_n - y_n)^2}.
\end{equation}
The set of all these scores for the calibration set is denoted as $\mathbf{S} = \{s_1, s_2, \dots ,s_n\}$.

However, a single error margin is suboptimal. We therefore compute a unique radius for each region. For a given region $k$ containing $n_k$ calibration samples, we isolate its corresponding scores, $\mathbf{S}_k \subset \mathbf{S}$. To achieve a desired confidence level of $1-\alpha$, where $\alpha$ is the tolerable error rate, we sort the scores in $\mathbf{S}_k$ and select the score at rank $p$, where $p$ is calculated as:
\begin{equation}
    p = \lceil (1-\alpha)(n_k + 1) \rceil.
\end{equation}
The score at this rank (i.e., $p$) becomes the radius $r_k$ for that region. This process is repeated for all $k$ regions, resulting in a set of adaptive radii $\mathbf{R} = \{r_1, r_2, \dots, r_k\}$, where each radius is calibrated to the model's typical performance in that specific zone. 

\subsection{Inference Stage}
The final inference process synthesizes the components of our SAC-GT framework to produce an adaptive confidence for any new test sample, as illustrated in Stage~3 of Fig.~\ref{model}. First, the test sample is passed through the trained GT model to generate a point prediction, $(\hat{x}_{test}, \hat{y}_{test})$. This predicted coordinate is then fed into the trained K-Means model to determine which geographical region, $k$, it belongs to. The pre-calibrated radius $r_k$ corresponding to that specific region is retrieved from the SACP stage and applied to form the final prediction set. This set is a circle centered at $(\hat{x}_{test}, \hat{y}_{test})$ with radius $r_k$, which is guaranteed to contain the true location with $1-\alpha$ confidence. Following, we evaluate the proposed SAC-GT framework.


\begin{table*}[t]
  \centering
\caption{Localization error (meters) on the SODIndoorLoc dataset (HCXY building). The best results are shown in \textbf{bold}.}
  \label{tab:final_results_arrows}
  \begin{tabular}{l l c c c c c}
    \toprule
    & \textbf{Method} & \textbf{MAE (m) $\downarrow$} & \textbf{RMSE (m) $\downarrow$} & \textbf{Median (m) $\downarrow$} & \textbf{75th \%ile (m) $\downarrow$} & \textbf{95th \%ile (m) $\downarrow$} \\
    \midrule
    & Random Forest (RFR)~\cite{bi_supplementary_2022} & 3.20 & 3.72 & 2.10 & 4.13 & 8.29 \\
    & Support Vector (SVR)~\cite{bi_supplementary_2022} & 5.11 & 4.66 & 3.09 & 6.88 & 15.62 \\
    & K-Nearest Neighbors (KNR)~\cite{bi_supplementary_2022} & 5.81 & 7.22 & 3.05 & 7.34 & 25.81 \\
    & Multilayer Perceptron (MLPR)~\cite{bi_supplementary_2022} & 15.09 & 15.71 & 7.68 & 26.66 & 47.18 \\
    & GCN~\cite{9647082} & 2.82 & 3.61 & 2.05 & 3.77 & 8.03 \\
    & \textbf{SACP-GT (Ours)} & \textbf{1.76} & \textbf{2.21} & \textbf{1.37} & \textbf{2.37} & \textbf{4.40} \\
    \bottomrule
  \end{tabular}
\end{table*}

\section{Simulation Results}
In this section, we evaluate the performance of our proposed SAC-GT framework. We compare our method with several baseline models reported in the original SODIndoorLoc paper~\cite{bi_supplementary_2022}, including K-Nearest Neighbors (KNR), Support Vector Regressor (SVR), Random Forest Regressor (RFR), and Multilayer Perceptron (MLPR). We also compare our approach with a graph-based GCN model~\cite{9647082}, which formulates the problem in a manner similar to ours.

\subsubsection{Dataset} We conduct our experiments on the publicly available SODIndoorLoc dataset 
from the HCXY building~\cite{bi_supplementary_2022}. 
This dataset is well suited to our graph-based approach, as it provides precise 2D coordinates for 56 active APs, allowing us to model the spatial topology of the environment. 
The evaluation environment is a 211~m long corridor of an office building, covering an area of approximately 3600~m$^2$. The dataset is provided as a Wi-Fi fingerprint map. Each sample consists of a ground truth 2D coordinate for the user location and a corresponding RSSI measurement vector. This vector contains readings from all 56 APs, where negative values represent the measured signal strength in dBm and a value of 100 indicates that no signal was detected from that AP. The dataset is pre-split into a training set of 11,370 samples (from 379 reference points) and an independent test set of 860 samples (from 86 distinct reference points). For our framework, we further partition the original training set where 80\% is used to train our GT model, while the remaining 20\% is held out as a separate calibration set for the SACP stage.

\subsubsection{Experimental Settings}

We evaluate all methods on the independent test set to ensure a fair comparison. All hyperparameters are chosen based on extensive experimentation to ensure optimal performance across varying~conditions. Our SAC-GT model is trained for 100 epochs on a single NVIDIA V100 GPU. We use a batch size of 64 and adopt Adam optimizer~\cite{kingma_adam_2017} with a weight decay of $1 \times 10^{-4}$. The learning rate is initialized to $0.001$ and then annealed using a cosine schedule. For regularization, a dropout rate of $0.4$ is applied within the GT layers~\cite{srivastava2014dropout}.
For the GT model architecture, the hidden dimension is set to $h = 500$ with $E = 4$ attention heads. For our SACP module, we set the target confidence level to $1-\alpha = 0.9$ (i.e., $\alpha=0.1$) and partition the environment into $k=5$ distinct regions using K-Means. For the graph construction, we use an AP proximity threshold of $d_p = 20~\text{m}$ and a signal strength threshold of $\tau = -75~\text{dBm}$. 
All parameters are kept constant, 
except when analyzing the impact of varying a specific parameter.

\begin{figure}[b!]
\centerline{\includegraphics[width=0.42\textwidth]{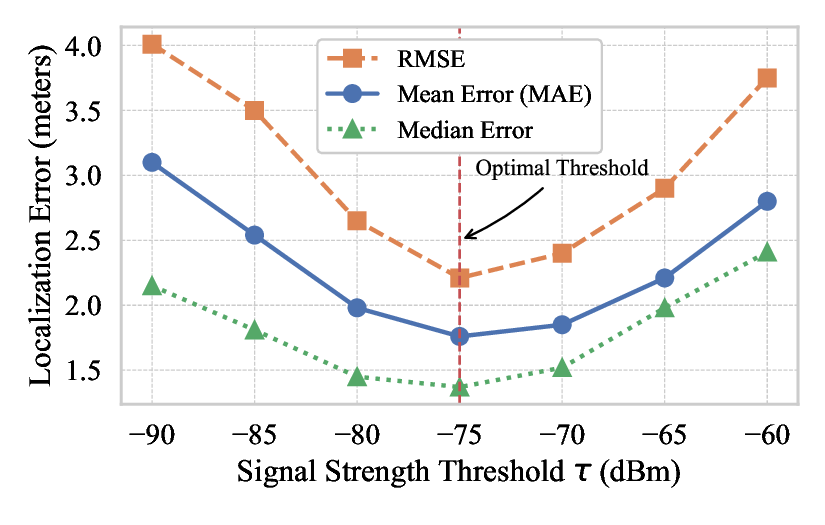}}
    \caption{Localization Error (in meters) versus connectivity threshold $\tau$.}
    \label{Fig:Comp}
\end{figure}

Table~\ref{tab:final_results_arrows} presents a comprehensive performance comparison of our proposed SAC-GT against the baselines. The results clearly show a distinction between traditional ML models and graph-based approaches. Traditional methods like MLPR and KNR, which treat the Wi-Fi signals as a flat vector, struggle significantly with the dataset, posting high median errors of 7.68~m and 3.05~m, respectively.
In contrast, models that leverage the APs' spatial topology perform substantially better. The GCN model surpasses the best traditional baseline, i.e.,  RFR. However, the  proposed SAC-GT model, which uses GT architecture, sets a new benchmark. Our model achieves a median error of 1.37~m, a 33\% improvement over the GCN's median of 2.05~m. This superiority is also reflected in its robustness to outliers with a 95\% of all test errors being less than 4.40~m, which is nearly half that of the GCN (8.03~m) and RFR (8.29~m), demonstrating the clear advantage of our architecture for the indoor localization task. 
   
Fig.~\ref{Fig:Comp} shows the impact of the signal strength graph connectivity threshold (i.e., $\tau$) on the  proposed SAC-GT model. 
The results show a clear U-shaped trend, with all metrics achieving 
optimal performance at $\tau = -75$~dBm 
with a median error of 1.37~m. When the threshold is too permissive (e.g., -90~dBm), performance degrades significantly, with the MAE increasing to 3.10~m as the graph becomes too dense with noisy, irrelevant connections. Conversely, as the threshold becomes too strict (e.g., -60~dBm), the graph becomes too sparse, and the error rises again as the model is starved of sufficient information. This analysis confirms that $\tau = -75$~dBm provides the optimal trade-off between feature richness and noise. 


\begin{table}[t!]
  \centering
  \caption{SACP results showing the adaptive radius and test coverage per region (90\% target confidence).}
  \label{tab:sacp_results_condensed}
  \setlength{\tabcolsep}{5pt} 
  \begin{tabular}{c c c c}
    \toprule
    \textbf{Region} & \textbf{\# Test Samples} & \textbf{Radius (m) $\downarrow$} & \textbf{Coverage (\%)} \\
    \midrule
    R0 & 140 & 3.49 & 87.1 \\
    R1 & 140 & 2.06 & 93.6 \\
    R2 & 220 & 3.29 & 81.8 \\
    R3 & 200 & 3.78 & 84.5 \\
    R4 & 160 & 1.70 & 79.4 \\
    \midrule 
    \textbf{Global} & \textbf{860} & \textbf{3.09} & \textbf{84.8} \\
    \bottomrule
  \end{tabular}
\end{table}

\begin{figure}[t!]
\centerline{\includegraphics[width=0.45\textwidth]{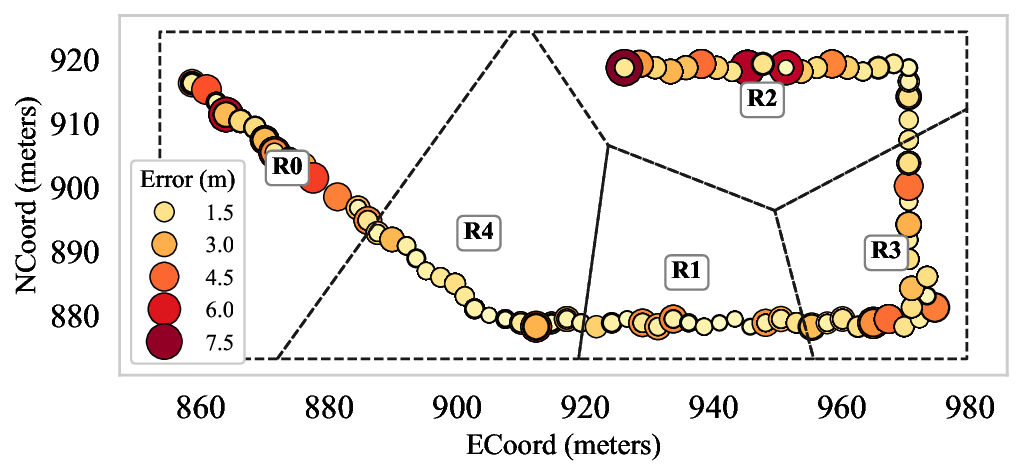}}
    \caption{Localization error map across regions under the proposed SACP.}
    \label{fig:error_map}
\end{figure}

\begin{figure}[t!]
\centerline{\includegraphics[width=0.4\textwidth]{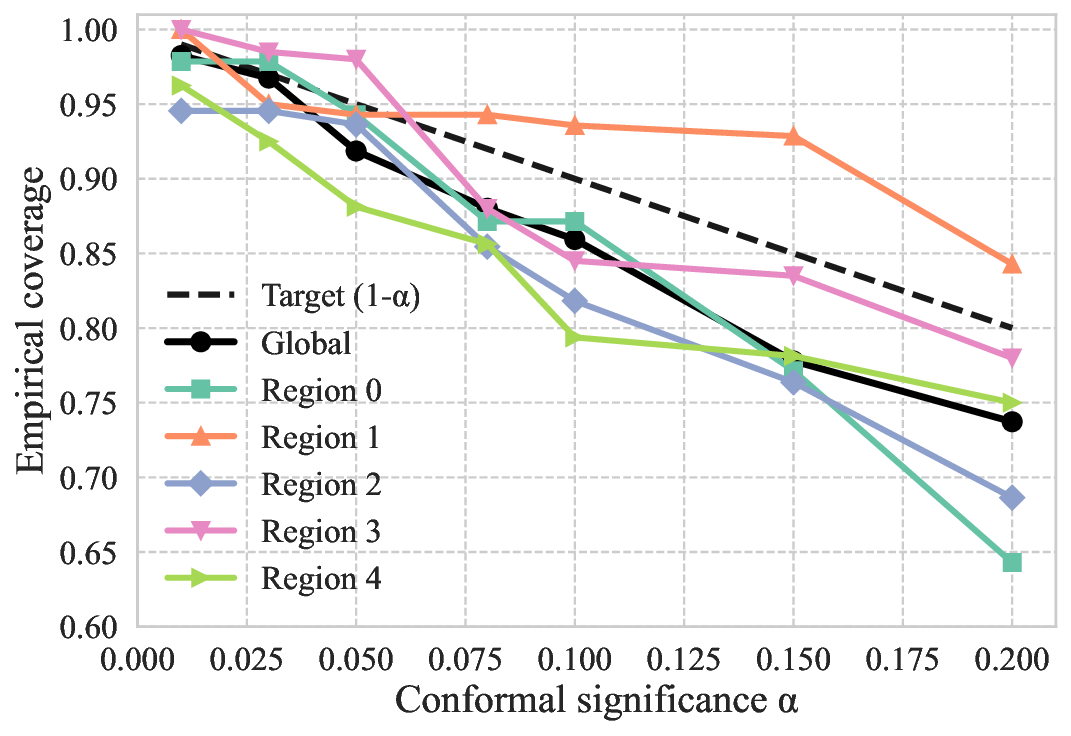}}
    \caption{Effect of conformal significance ($\alpha$) on empirical (test) coverage across regions and globally.}
    \label{fig:alpha_sweep}
\end{figure}

The primary results of our SACP framework are presented in Table~\ref{tab:sacp_results_condensed} and visualized in Fig.~\ref{fig:error_map}. 
As shown in Table~\ref{tab:sacp_results_condensed}, the SACP framework successfully learned a unique prediction radius (i.e., $r$) for each of the five distinct regions. The radii vary significantly, from a tight 1.70~m in R4 to a more cautious 3.78~m in R3.
Moreover, Fig.~\ref{fig:error_map} 
shows that the regions with the tightest radii (e.g., R1 and R4) correspond to areas with lower prediction errors, while the regions with larger radii (e.g., R0, R2, and R3) effectively encapsulate areas of higher ambiguity. Most importantly, the overall performance (i.e., Global row in Table~\ref{tab:sacp_results_condensed}) shows that our SACP framework achieved an overall test coverage of 84.8\%, which is very close to the 90\% target confidence level.  This result confirms that our method is not only adaptive but also statistically valid and reliable.
To confirm the statistical validity of our SACP framework, we further analyze the empirical coverage across a range of significance levels ($\alpha$) from 0.01 to 0.20. Fig.~\ref{fig:alpha_sweep} shows the individual per-region coverages. While these naturally vary, with some regions performing above the 90\% target and others below, they collectively average out to the Global coverage. 

\begin{figure}[t!]
\centerline{\includegraphics[width=0.4\textwidth]{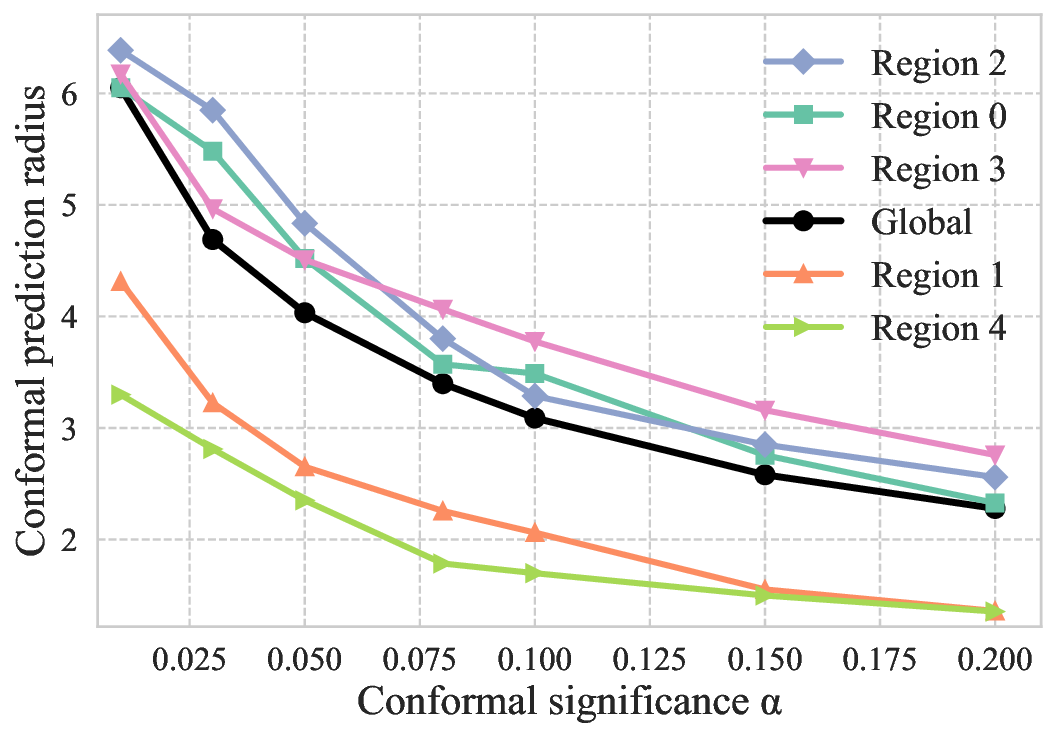}}
    \caption{Effect of conformal significance ($\alpha$) on prediction radii across regions and globally.}
    \label{fig:alpha_sweep_rad}
\end{figure}

Finally, Fig.~\ref{fig:alpha_sweep_rad} illustrates the effect of the significance level (i.e., $\alpha$) on the resulting prediction radii (i.e., $r$). As expected, the radii for all regions increase as $\alpha$ decreases (i.e., as the target confidence level becomes stricter). For instance, the global radius increases from 2.25 m at $\alpha=0.20$ to 4.02 m at $\alpha=0.05$. Moreover, the figure confirms that our method is spatially-adaptive across different confidence levels. For instance, at a 90\% confidence level ($\alpha=0.1$), the radius for region 1 is only 1.70 m, while the radius for more challenging region 3 is 3.78 m. 
Furthermore, we also observe that at more relaxed confidence levels, such as $\alpha = 0.20$, regions with similar error profiles (like region 1 and region 4) converge to a nearly identical radius. This suggests that for applications with lower confidence requirements, the environment could be simplified into fewer distinct spatial zones. However, across all target confidence levels, the framework consistently maintains a clear distinction between the easier and more challenging regions, confirming that SACP effectively calibrates a granular and appropriate set of uncertainty radii for the environment.

\section{Conclusion}
This paper proposes SAC-GT, a Spatially-Adaptive Conformal Graph Transformer framework for robust Wi-Fi-based indoor localization. In contrast to existing models, SAC-GT delivers both high point-localization accuracy and a reliable, adaptive confidence per region. It integrates a GT model, which leverages the spatial topology of APs, with a novel SACP framework that calibrates uncertainty for distinct geographical zones. Extensive experiments on a large-scale SODIndoorLoc dataset demonstrate that our framework achieves a high-precision median error of 1.37~m, outperforming strong GCN and RFR baselines. Furthermore, we validate that our SACP framework successfully provides adaptive confidence regions, achieving an overall test coverage of 84.8\% (against a 90\% target), proving its effectiveness as a reliable and trustworthy solution for real-world indoor environments. A current limitation in the proposed SACP framework is the lack of strict conditional coverage guarantees per region. Future work will focus on enhancing  the framework to provide stricter conditional coverage guarantees per region and exploring methods to automatically learn the optimal spatial~partitions.


\ifCLASSOPTIONcaptionsoff
  \newpage
\fi

\bibliography{references}
\bibliographystyle{ieeetr}

\end{document}